\documentclass[conference,10pt,letterpaper]{IEEEtran}
\usepackage{calfa-style}
\title{CALFA-KV: Joint Calibration and Fisher-Guided Anchors for 1-Bit KV Cache Compression}

\author{
\IEEEauthorblockN{
Xiancai Tian,
Leichao Zhang\textsuperscript{*},
Shenhui Zhang\textsuperscript{\dag},
Yi Shan\textsuperscript{\dag},
Hailong Li,
Feng Lin,
Xiaotian Yu,
Aijun Li
}
\IEEEauthorblockA{
Shenzhen Intellifusion Technologies Co. Ltd., Shenzhen, China\\
\texttt{\{tian.xiancai, zhang.leichao, zhang.shenhui, shan.yi,}\\
\texttt{li.hailong, lin.feng, yu.xiaotian, li.aijun\}@intellif.com}
}
\IEEEauthorblockA{
\footnotesize
\textsuperscript{*}Corresponding author.
\quad
\textsuperscript{\dag}Equal contribution.
}
}

\date{}
\usepackage{booktabs}
\usepackage{graphicx}
\usepackage[table]{xcolor}
\begin{document}
\maketitle
\thispagestyle{plain}
\begin{abstract}

1-bit key–value (KV) cache compression can combine quantizer calibration with selective high-precision storage to reduce reconstruction error. We propose Calibration and Fisher-guided Anchors for KV cache compression (CALFA-KV). Offline, we jointly fit a key center, radial normalization parameters, and K/V vector codebooks to match the frozen model's projected attention output. We also estimate channel-wise Fisher-type sensitivities from squared gradients of a reference next-token cross-entropy loss. Online, we use these sensitivities to weight the squared reconstruction errors of each KV state. Here, a KV state is one token's key–value pair within one layer and KV head. We retain the highest-scoring states at high precision under a per-layer, per-KV-head budget. Under exact restoration, we bound the excess risk of score-based anchor selection relative to the optimum at the same anchor budget, where risk is the conditional mean square of the first-order loss change. With a budget of 56 additional FP16 anchors per layer and KV head, CALFA-KV (d8b8) outperforms all evaluated nominal 1-bit baselines on the mean score across seven benchmarks for each of the three models. At the same anchor budget, CALFA-KV (d8b10) uses a nominal index rate of 1.25 bits per channel and narrows the gap to BF16 to 0.90–2.01 points on the same metric.

\end{abstract}
\section{Introduction}
\label{sec:1}

For 1-bit key–value (KV) cache compression, key errors alter attention weights, and value errors alter their weighted sum. Here, a KV state is one token's key–value pair within one layer and KV head. A fixed-rate quantizer assigns the same index cost to each state, yet the remaining reconstruction error can be concentrated in a small number of states.

Existing work improves the quantization representation and sensitivity-aware fitting \citep{kvquant,tasq}, or allocates additional precision to important tokens \citep{mikv,zipcache,antkv}. Calibration changes the remaining reconstruction errors, while high-precision selection determines which KV states to retain at higher precision.

Quantization errors vary across KV states, and errors of similar magnitude can affect model behavior differently. Under a limited high-precision storage budget, the challenge is to identify which states are most beneficial to preserve. This requires assessing the consequences of quantization errors beyond their reconstruction magnitude alone.

We present CALFA-KV, short for \textbf{CAL}ibration and \textbf{F}isher-guided \textbf{A}nchors.\footnote{Code is available at \url{https://github.com/XiancaiTian/CALFA-KV}.} Offline, we jointly fit the key center, radial normalization parameters, and K/V codebooks to match each layer's projected attention output. We also estimate channelwise Fisher-type sensitivities offline from squared gradients of a reference next-token cross-entropy loss. With the quantizer frozen, we weight each KV state's realized reconstruction errors by these sensitivities in matching scoring coordinates: normalized key coordinates with the stored scale held fixed, and the original value coordinates. During inference, we retain up to $N$ of the highest-scoring KV states at high precision for each KV head in each layer. We keep a quantized representation of every unprotected historical state.

Building on prior work, CALFA-KV links joint quantizer calibration to anchor selection through the reconstruction errors of the calibrated quantizer. We make three contributions:

\begin{itemize}
\item We jointly fit the pre-RoPE key center, radial scale parameters, and K/V codebooks to match each layer’s projected attention output. We compute each candidate update with the current codeword indices fixed. We then use the candidate parameters to re-encode the K/V states and recompute the projected attention outputs with the deployed quantizer and attention kernel. We accept the update only if it reduces the output error on the fitting set and passes the validation checks.
\item After freezing the quantizer, we estimate per-channel Fisher-type sensitivities offline from squared gradients of the reference model's next-token cross-entropy loss, using the same coordinates as the K/V reconstruction errors. During inference, we use these sensitivities to weight the squared reconstruction errors of each KV state and maintain a streaming top-$N$ set per layer and KV head.
\item We evaluate CALFA-KV on three language models across seven benchmarks covering mathematical reasoning, instruction following, code generation, long-context understanding, and scientific question answering. CALFA-KV (d8b10) uses a nominal index rate of 1.25 bits per channel. With a separate budget of 56 additional FP16 anchors per layer and KV head, its seven-benchmark mean scores are 0.90–2.01 points below BF16 across the three models.
\end{itemize}

The remainder of this paper is organized as follows. Section 2 reviews related work. Section 3 presents CALFA-KV's quantizer, joint calibration, and Fisher-guided anchor selection. Section 4 describes the evaluation setup, accuracy comparisons, and ablation study. Section 5 concludes the paper. Appendix A analyzes anchor selection, Appendix B explains the scoring coordinates and attention updates, and Appendix C details calibration and deployment.

\section{Literature Review}
\label{sec:2}

We review how prior work represents low-bit KV states, calibrates the quantizer, and allocates additional precision during inference.

\subsection{Low-Bit KV Representations}
\label{sec:2.1}

Low-bit KV quantizers differ in what they encode together. KIVI quantizes keys per channel and values per token \citep{kivi}. KVQuant uses non-uniform levels for pre-RoPE keys and stores sparse outliers at higher precision \citep{kvquant}. Coupled Quantization (CQ) uses vector quantization (VQ) to encode groups of channels with vector codewords \citep{cq}.

Transforming the KV distribution can make it easier to quantize. NSNQuant uses normalization, centering, and a Hadamard transform to enable a calibration-free shared codebook, while VecInfer pairs key smoothing and a Hadamard transform with VQ lookup attention \citep{nsnquant,vecinfer}. OSCAR derives rotations from attention-aware covariance targets for scalar INT2 quantization, and NOVA-KV combines key mean centering and attention-preserving transforms with fixed-rate VQ \citep{oscar,novakv}. QuaRot applies computationally invariant rotations in a broader model-quantization pipeline \citep{quarot}.

TaSQ provides the closest representation-level reference through query-guided channel weighting, a token scale shared across KV heads, and covariance-aware grouping \citep{tasq}. CALFA-KV starts from TaSQ-style covariance grouping and refines the key-channel grouping with codebooks refitted at the target rate.

\subsection{Quantizer Calibration Objectives}
\label{sec:2.2}

Reconstruction-based calibration minimizes the error between the reconstructed and reference K/V states. KVQuant and CQ optimize quantization levels or codebooks using sensitivity-weighted reconstruction errors. \citep{kvquant,cq}. Other methods measure error further downstream. SKVQ tunes clipping with attention-output mean squared error, and SpinQuant learns rotations against the quantized model's loss \citep{skvq,spinquant}. For multi-head latent attention (MLA), RoLaQ calibrates rotations and ranges jointly, while RAQ-MLA uses different error metrics for the shared KV latent and positional-key cache \citep{rolaq,raqmla}.

OptR centers keys after RoPE and first fits key rotations using attention-distribution KL divergence and post-projection output error. It then fits value rotations under the selected quantized-key path \citep{optr}. CALFA-KV jointly fits the pre-RoPE key center, radial scale parameters, and K/V codebooks against a shared layerwise projected attention-output objective. The channel weights and refined key groups stay fixed during this joint calibration.

\subsection{Selective High-Precision KV Retention}
\label{sec:2.3}

Precision allocation can operate at different granularities. MiKV and ZipCache vary precision with token importance \citep{mikv,zipcache}. FGMP scores weight and activation blocks using sensitivity-weighted quantization perturbations and applies an offline-calibrated threshold to activation blocks during inference \citep{fgmp}. MixKVQ ranks key channels using query magnitudes and a quantization-step proxy \citep{mixkvq}. AnTKV learns codebooks offline and selects high-precision tokens with an online score based on attention weights and, for keys, query norms. It uses a sliding-window approximation during decoding \citep{antkv}.

Retained information can also help encode other states. AnchorKV uses high-precision anchors as directions for reconstructing other states and allocates quantized residuals by attention-output utility \citep{anchorkv}. GEAR adds low-rank and sparse components, AQUA-KV codes prediction residuals, and the inner-product variant of TurboQuant adds a residual code \citep{gear,aquakv,turboquant}.

H2O, StreamingLLM, and SnapKV retain selected historical KV states, whereas ReadKV adapts the number of bits fetched for each query \citep{h2o,streamingllm,snapkv,readkv}. CALFA-KV maintains a low-bit representation independently of anchor selection and augments it with FP16 copies of selected K/V pairs. With the quantizer fixed, it scores each quantized historical state using its realized squared K/V reconstruction errors, weighted by offline-calibrated Fisher-type sensitivities, and retains the highest-scoring states as anchors, with a budget of $N$ per layer and KV head for each request.

\section{Methodology}
\label{sec:3}

\begin{figure*}[t]
\centering
\includegraphics[width=\textwidth]{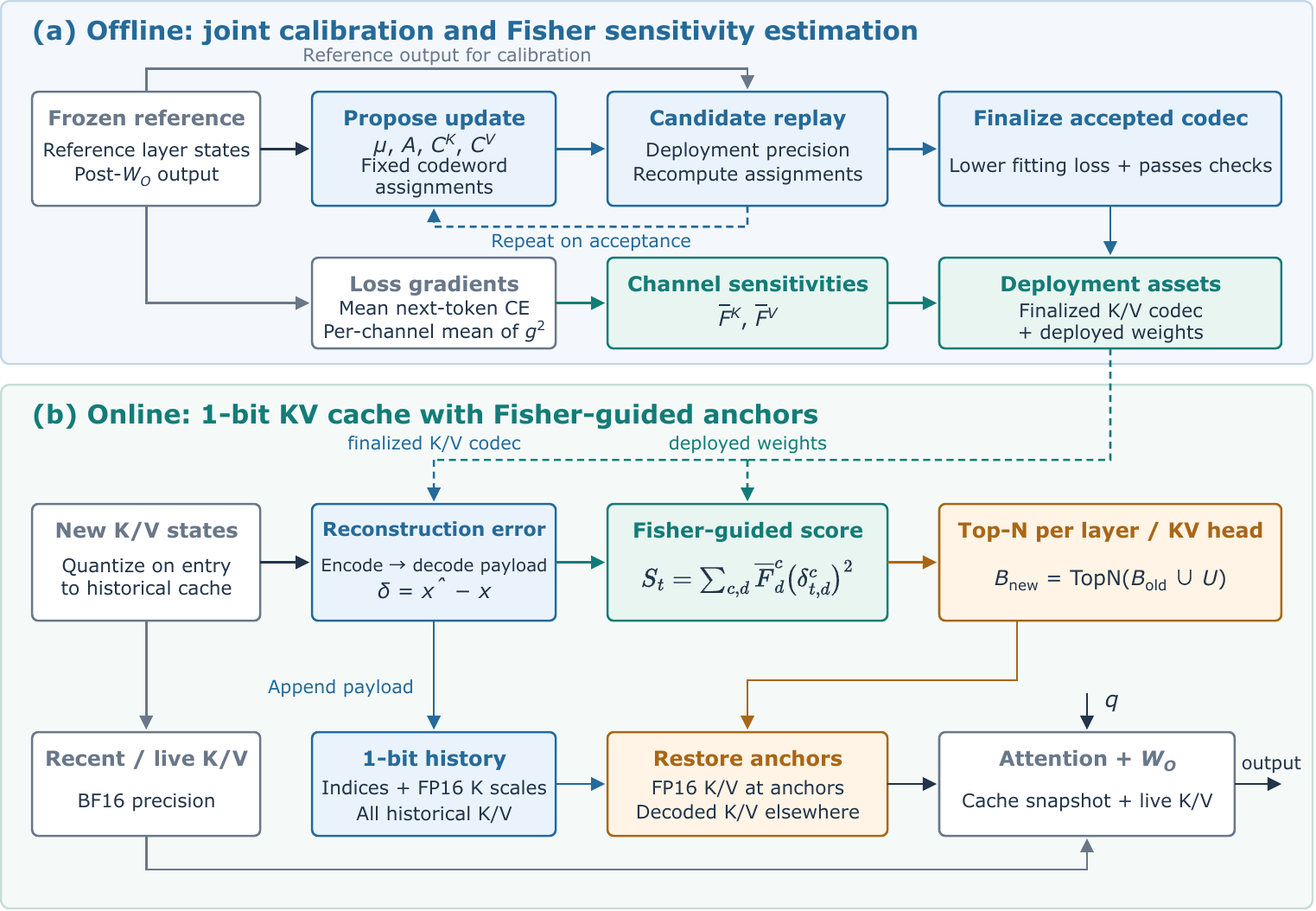}
\caption{CALFA-KV for 1-bit KV cache compression. (a) Offline joint calibration proposes updates with fixed codeword assignments and tests them using deployment-precision replay after re-encoding. Channel sensitivities are estimated separately from reference-loss gradients. (b) During inference, sensitivity-weighted reconstruction errors rank historical KV states for FP16 retention, with up to $N$ anchors per layer and KV head. Attention combines decoded history, restored anchors, and unquantized recent and live K/V. The nominal 1-bit rate refers to codeword indices.}
\label{fig:1}
\end{figure*}

CALFA-KV links offline calibration of a low-bit K/V codec (encoder and decoder) to online anchor selection (Figure~\ref{fig:1}). Offline, we jointly calibrate the codec to match the frozen model's projected attention output and separately estimate channel sensitivities from squared reference-loss gradients. Online, we weight each eligible KV state's squared reconstruction error by these sensitivities and retain up to $N$ of the highest-scoring states as high-precision anchors for each layer and KV head. The remaining eligible states use their quantized representations.

\subsection{Key and value representation}
\label{sec:3.1}

We represent historical keys and values using groupwise vector quantization. Keys are quantized before RoPE so that codebooks need not represent the additional position-dependent rotations. During attention computation, quantized keys are reconstructed and rotated with RoPE at their original token positions. Values are quantized directly in the coordinates used by attention. For each layer, we learn a key center $\mu$, a positive diagonal radial matrix $A$, and separate K/V codebooks. A fixed diagonal matrix $D$ weights key-channel distances, and a fixed within-head permutation $P$ assigns key coordinates to groups while keeping RoPE-coupled units intact so that each codeword contains both components needed for their rotation.

For a layer with $H$ KV heads of dimensions $d_K,d_V$, let $k_t\in\mathbb R^{D_K}$ and $v_t\in\mathbb R^{D_V}$ concatenate the token-$t$ key and value vectors, with keys taken after any model-native key normalization and before RoPE, where $D_K=Hd_K$ and $D_V=Hd_V$. A group width $g$ divides both head dimensions, giving $G_K=d_K/g$ key groups and $G_V=d_V/g$ value groups per head. Each group uses a codebook containing $K_c$ codewords, each a vector in $\mathbb R^g$. The deployed codewords are stored in FP16. Each token stores one FP16 key scale $n_t$, shared across the layer's KV heads. We impose a positive lower bound \(\epsilon\) on the key scale. The encoding is then defined as follows.

\[
\begin{aligned}
n_t&=\operatorname{FP16}\!\left[\max\!\left(\epsilon,\frac{\|A(k_t-\mu)\|_2}{\sqrt{D_K}}\right)\right],\\
z_t&=\frac{PD(k_t-\mu)}{\operatorname{FP32}(n_t)}.
\end{aligned}
\tag{1}
\label{eq:1}
\]

The center removes a persistent offset, $A$ controls the shared token scale, and $D$ weights channel errors when choosing codewords. The shared scale reduces token-to-token magnitude variation before codeword lookup. We initialize $A=D$ and calibrate $A$ while keeping the distance weights $D$ and the grouping $P$ fixed. We tie the ratios $A_{jj}/D_{jj}$ across channels within each RoPE-coupled unit, so these channels share one learnable multiplier relative to $D$. To keep calibration consistent with deployment, codebook fitting, calibration acceptance, cache writes, and inference follow the same scale computation: each token's key scale is rounded to FP16 and then promoted to FP32 for encoding and decoding. The stored FP16 value must be finite and strictly positive. Appendix C.1 details this parameterization and the fixed FP32 summation order used to compute the scale.

Key groups come from $z_t$. Value groups use consecutive channels in the original value coordinates and unit scale. For head $h$, group $b$, and the separate K/V codebooks, the stored indices select

\[
\begin{aligned}
i^K_{t,h,b}&=\argmin_{0\le j<K_c}\|z_{t,h,b}-C^K_{h,b,j}\|_2^2,\\
i^V_{t,h,b}&=\argmin_{0\le j<K_c}\|v_{t,h,b}-C^V_{h,b,j}\|_2^2.
\end{aligned}
\tag{2}
\label{eq:2}
\]

Each index replaces $g$ channel values with a codeword identifier. Concatenating the selected codewords gives $\widehat z_t$ and $\widehat v_t$. In real arithmetic, the reconstruction is

\[
\begin{aligned}
\widehat k^{\mathrm{pre}}_t
&=\mu+D^{-1}P^{\mathsf T}
\!\left(\operatorname{FP32}(n_t)\widehat z_t\right),\\
\widehat k^{\mathrm{post}}_t&=R_t\widehat k^{\mathrm{pre}}_t,\\
\widehat v_t&=\operatorname{Concat}_{h,b}C^V_{h,b,i^V_{t,h,b}}.
\end{aligned}
\tag{3}
\label{eq:3}
\]

Here $R_t=\operatorname{blkdiag}(R_{t,1},\ldots,R_{t,H})$, where $R_{t,h}\in\mathbb R^{d_K\times d_K}$ applies RoPE to KV head $h$ at token position $t$. Equation~\eqref{eq:3} describes reconstruction in real arithmetic. The deployed attention kernel rounds dot-product operands to FP16, applies the stored token scale after the key dot product, and adds the contribution from the rotated center separately. Calibration acceptance uses this deployed computation (Appendix C.1).

Anchor scoring uses the deployed K/V reconstructions. Reconstructed keys are mapped to pre-RoPE coordinates using the inverse transform defined by the model's cached sine and cosine values. Key errors are then measured in normalized encoding coordinates with each token's stored scale held fixed, while value errors are measured directly in the original value coordinates (Section~\ref{sec:3.3}).

Before joint calibration, we construct $D$ from calibration data and then determine $P$; both remain fixed during joint calibration. We adopt TaSQ's query-guided channel weighting and covariance-aware hierarchical grouping to initialize $D$ and $P$ \citep{tasq}.

For KV head $h$, let $\mathcal A_h$ be the set of query heads that share its K/V states. For calibration sequence $i$ and query head $a\in\mathcal A_h$, let $q_{i,s,a}$ be the query at position $s$ after any model-native query normalization and before RoPE, and define $\bar q_{i,s,a}=R_{s,h}q_{i,s,a}$. Let $\mathcal T_h$ contain all valid causal tuples $(i,s,t,a)$ from the calibration sequences, with $t\le s$ and $a\in\mathcal A_h$. The query second moment aligned to the pre-RoPE key coordinates and the channel weights are

\[
\begin{aligned}
M_h
&=\frac{1}{|\mathcal T_h|}\sum_{(i,s,t,a)\in\mathcal T_h}
R_{t,h}^{\mathsf T}\bar q_{i,s,a}\bar q_{i,s,a}^{\mathsf T}R_{t,h},\\
m_{h,j}
&=[M_h]_{jj},\\
\eta_h
&=\max\!\left(\epsilon_{\mathrm{abs}},
\epsilon_{\mathrm{rel}}\max_j m_{h,j}\right),\\
w_{h,j}
&=\sqrt{\frac{\max(m_{h,j},\eta_h)}
{d_K^{-1}\sum_{j'=1}^{d_K}\max(m_{h,j'},\eta_h)}}.
\end{aligned}
\tag{4}
\label{eq:4}
\]

The absolute and relative floors $\epsilon_{\mathrm{abs}}>0$ and $\epsilon_{\mathrm{rel}}>0$ stabilize the channel-energy estimates and ensure positive weights. Within each head, normalization gives $d_K^{-1}\sum_jw_{h,j}^2=1$. We concatenate $w_{h,j}$ into $D=\operatorname{diag}(w)$, round it to FP32, and freeze it. Using only the diagonal of $M_h$, this scaling emphasizes channels with higher aligned query energy and provides a proxy for squared attention-logit error.

To construct $P$, we set $\mu=0$ and $A=D$ and form normalized keys $\xi_t=Dk_t/\operatorname{FP32}(n_t)$ using Eq.~\eqref{eq:1}. For each head, let $\Sigma_h$ be their covariance and $r_h$ the common width of a RoPE-coupled coordinate unit. We keep each unit intact and require $g/r_h=2^{s_h}$ for an integer $s_h\ge0$. At each of the $s_h$ hierarchy levels, we merge pairs of units using the covariance-volume cost

\[
\begin{aligned}
\gamma_h(I)
&=\det\!\left(\Sigma_h[I,I]+\lambda_h I_{|I|}\right)^{1/|I|},\\
\mathcal M_h^\star
&=\argmin_{\mathcal M\in\operatorname{PM}(\mathcal U_h)}
\sum_{(U,V)\in\mathcal M}\gamma_h(U\cup V),
\end{aligned}
\tag{5}
\label{eq:5}
\]

Here $\lambda_h>0$, $I_{|I|}$ is the identity matrix, $\mathcal U_h$ is the current set of coordinate units, and $\operatorname{PM}$ denotes perfect matchings. The hierarchy produces an initial $P_h$ \citep{tasq}, and $P=\operatorname{blkdiag}(P_1,\ldots,P_H)$. Its covariance-volume cost is a heuristic for grouping correlated channels; finite codebook capacity is assessed by fitting codebooks at the target rate. The construction then uses two stages: reconstruction-based regrouping with attention-output guards, followed by pair-swap refinement accepted by attention-distribution KL. In the pair-swap stage, a proposal and a control retaining the current $P$ use the same calibration samples, initialization rule, random seeds, and maximum fitting budget. Acceptance requires KL improvement over both the current result and the refitted control, together with non-degradation guards on stratified KL and attention-output error. Appendix C.1 gives the distinct criteria for both stages. The final $D$, $P$, and group structure remain fixed during Section~\ref{sec:3.2}.

In the configuration labels, \texttt{d} denotes the dimension $g$ of each group, and \texttt{b} denotes the number of bits per codeword index. The index rate is $\log_2 K_c/g$ bits per channel. With $g=8$, the d8b8 and d8b10 configurations use $K_c=256$ and $K_c=1024$, respectively, giving index rates of 1.00 and 1.25 bits per channel. Scales, codebooks, and high-precision KV states contribute additional storage. Appendix C.1 distinguishes these representation parameters from the configuration coverage of the accompanying implementation.

\subsection{Joint calibration against attention output}
\label{sec:3.2}

The coordinate-construction stage above provides frozen $D/P$ and codebooks fitted to their current groups with the deployment encoder. Before updating the center and radial parameters, we jointly warm-start the K/V codebooks against projected attention-output error with $\mu_0=0$ and $A_0=D$. We then optimize $\theta_l=(\mu_l,A_l,C^K_l,C^V_l)$ while keeping $D$, $P$, the group structure, and the model weights fixed. This common objective is needed because vector reconstruction alone misses how key errors change softmax weights and value errors change the weighted sum. The two-stage design therefore separates discrete finite-rate coordinate construction from continuous calibration of the representation that deployment actually reads.

Each layer is calibrated separately from frozen reference states, with anchors disabled. The replay quantizes historical KV states and retains the recent and live branches at reference precision, using the native causal mask, cache mask, grouped-query mapping, and output projection. For calibration window $w$, let $o^\star_{l,w}$ concatenate the reference outputs after $W_O$ at its fixed evaluated query positions, including all projected channels. Define $o^{\mathrm{nat}}_{l,w}(\theta_l)$ by the same concatenation for the replay with candidate parameters $\theta_l$. Each norm below is therefore a vector norm over these selected query outputs, rather than all token positions in the window. Appendix C.1 specifies the query positions and Fit/Select split. For finite reference energy $0<E_l<\infty$, we minimize normalized squared output error over fitting windows $\mathcal F$:

\[
\begin{aligned}
\mathcal L_l(\theta_l)&=\frac{1}{2E_l}\sum_{w\in\mathcal F}
\|o^{\mathrm{nat}}_{l,w}(\theta_l)-o^\star_{l,w}\|_2^2,\\
E_l&=\sum_{w\in\mathcal F}\|o^\star_{l,w}\|_2^2.
\end{aligned}
\tag{6}
\label{eq:6}
\]

This objective measures the combined effect of key and value errors after attention and output projection, where the model actually uses the cached KV states. At each optimization step, assignments from the current stored representation are held fixed while the smooth replay and locally aligned scale surrogate in Appendix C.1 provide a continuous Gauss--Newton search direction. A trial is then materialized at deployment precision: parameters and codebooks are cast to their stored dtypes, Eq.~\eqref{eq:1} recomputes the deterministic FP16 scales, and K/V states are re-encoded using hard nearest-neighbor assignments within each group. Native replay evaluates the trial with the same masks, high-precision branches, and $W_O$ as deployment. The trial is accepted only if this hard replay strictly improves the objective on fitting views and passes non-degradation guards on separate selection views. Rejected candidates do not update the accepted state. The selection views participate in model selection and are not reported as independent test data. Thus continuous derivatives propose candidates, whereas the actual stored representation and native attention path decide acceptance. Appendix C gives the parameterization, deterministic reduction, and update rule.

\subsection{Fisher-guided anchor selection}
\label{sec:3.3}

Even after joint calibration, some KV states have costly reconstruction errors. We estimate channel sensitivities from the reference model's mean shifted next-token cross-entropy $\ell_i$ for sequence $i$. Let $D_{l,h}$ and $P_{l,h}$ be the head blocks of the frozen channel weights and permutation. For each captured position, we transform the signed pre-RoPE key gradient into the normalized encoding coordinates before squaring; the value gradient stays in its original coordinates:

\[
\begin{aligned}
g^K_{i,l,h,t}
&=\operatorname{FP32}(n_{i,l,t})P_{l,h}D_{l,h}^{-1}
\nabla_{k^{\mathrm{pre}}_{i,l,h,t}}\ell_i,\\
g^V_{i,l,h,t}&=\nabla_{v_{i,l,h,t}}\ell_i,\\
\widetilde F^c_{l,h,d}
&=\frac{\sum_i\sum_{t\in\mathcal M_i}(g^c_{i,l,h,t,d})^2}
{\sum_i|\mathcal M_i|},\quad c\in\{K,V\}.
\end{aligned}
\tag{7}
\label{eq:7}
\]

Each captured calibration position contributes once. Its stored key scale is treated as fixed side information during the coordinate transformation; we do not differentiate through the input-dependent normalization or FP16 rounding. Appendix B defines the loss positions, captured-state mask $\mathcal M_i$, and pooling. This loss is separate from the output-calibration objective in Eq.~\eqref{eq:6}. Clipping, shrinkage, and normalization give deployed weights $\overline F^c$ in the same scoring coordinates as the errors below.

When a historical KV state becomes eligible for anchor selection, we reconstruct its actual deployed post-RoPE key $\widetilde k^{\mathrm{post}}_{l,h,t}$ and value $\widetilde v_{l,h,t}$. Let $R^{\mathrm{cache}}_{t,h}$ denote the positional transform represented by the model's cached sine and cosine values. We recover $\check k^{\mathrm{pre}}_{l,h,t}=(R^{\mathrm{cache}}_{t,h})^{-1}\widetilde k^{\mathrm{post}}_{l,h,t}$ and form the errors and joint score:

\[
\begin{aligned}
\delta^K_{l,h,t}
&=\frac{P_{l,h}D_{l,h}(\check k^{\mathrm{pre}}_{l,h,t}-k^{\mathrm{pre}}_{l,h,t})}
{\operatorname{FP32}(n_{l,t})},\\
\delta^V_{l,h,t}&=\widetilde v_{l,h,t}-v_{l,h,t},\\
S_{l,h,t}
&=\sum_d\overline F^K_{l,h,d}(\delta^K_{l,h,t,d})^2\\
&\quad+\sum_d\overline F^V_{l,h,d}(\delta^V_{l,h,t,d})^2.
\end{aligned}
\tag{8}
\label{eq:8}
\]

The center cancels in the key-error difference, and the stored scale is consumed once. The inverse positional transform includes the finite-precision sine/cosine norm; the errors therefore retain the deployed reconstruction and rounding effects. A sensitive channel contributes little when its reconstruction is accurate, and a large error contributes little when its weight is small. The two sides contribute to one score for the paired K/V state.

Each anchor stores paired K/V values in FP16. With equal storage cost per KV state and a budget of at most $N$ anchors for each KV head, we retain the highest-scoring eligible KV states up to that budget. Appendix A proves this rule for the stated additive objective in the scoring coordinates. Exact restoration is an analysis condition: saving a KV state in a different dtype does not by itself establish zero error. If storage introduces restoration error, Equation~\eqref{eq:A11} gives the corresponding net proxy benefit.

Appendix A also compares the proxy with the conditional expected square of the linearized loss change in these fixed coordinates. If they differ by at most $\varepsilon_x$ for every feasible anchor set, the selected set has risk at most $2\varepsilon_x$ above the optimum for the same frozen states and budgets. Equation~\eqref{eq:A7} accounts for moment shift, omitted correlations, and deployed-weight differences. These terms are not estimated in the reported experiments.

\subsection{Streaming anchor maintenance}
\label{sec:3.4}

For each request, the selector maintains at most $N$ anchors per layer and KV head. A KV state becomes eligible for selection when it first enters the quantized history, either through direct quantization during initial or extended prefill, or through migration from the high-precision recent window. While its original K/V remain available, we compute its score from the deployed payload reconstruction using Equation~\eqref{eq:8}. Given the existing anchor set $B_{\mathrm{old}}$ and newly scored candidates $U$, we update

\[
B_{\mathrm{new}}=\operatorname{TopN}(B_{\mathrm{old}}\cup U).
\tag{9}
\label{eq:9}
\]

The codec and sensitivity weights remain fixed within each request, so each state's score remains fixed after admission. With a fixed position-based tie rule, this update maintains the highest-scoring eligible states under the budget (Appendix~\ref{sec:A.4}). New anchors store paired K/V values in FP16. States removed from the anchor set retain their quantized payloads and subsequently use those representations.

Attention uses the FP16 K/V copies for anchors and decoded payloads for other quantized historical states, together with recent and live K/V at reference precision. During prefill/extend, current live K/V are supplied directly to attention, and their cache representations are written afterward. During decode, the current token's K/V are first written to a high-precision slot and then read by attention. Restoring an anchor replaces both its key contribution to the attention logits and its value contribution to the weighted sum (Appendix~\ref{sec:B}).

Recent-to-history migration occurs between attention operations. Migrating states are encoded and scored while their original recent K/V remain available. Cache writes, anchor updates, and buffer reuse are synchronized so that active readers observe stable contents and mappings. Appendix~\ref{sec:C} specifies the flush schedule, admission procedure, and position/slot bookkeeping.

\section{Experiments}
\label{sec:4}

\subsection{Evaluation protocol}
\label{sec:4.1}

We report evaluations of Qwen3-4B-Instruct-2507, Llama-3.1-8B-Instruct, and Qwen3-30B-A3B-Instruct-2507. The evaluation covers seven benchmarks: GSM8K \citep{gsm8k}, IFEval \citep{ifeval}, HumanEval+ \citep{evalplus}, LiveCodeBench V5+V6 \citep{livecodebench}, LongBench \citep{longbench}, MATH500 \citep{math500}, and GPQA-Diamond \citep{gpqa}.These benchmarks contain a total of 7,814 questions. Table 1 summarizes the question counts, scoring metrics, and output limits.

\begin{table*}[t]
\centering\small
\setlength{\tabcolsep}{3pt}
\caption{Benchmark subsets and evaluation settings.}
\label{tab:2}
\begin{tabularx}{\textwidth}{@{}L{0.22\textwidth}L{0.13\textwidth}YL{0.18\textwidth}@{}}
\toprule
Subset & Questions & Primary score & Output limit (tokens) \\
\midrule
GSM8K & 1,319 & Numeric exact match with flexible answer extraction & 8,192 \\
IFEval & 541 & Strict prompt-level accuracy: all constraints in a prompt must pass & 1,280 \\
HumanEval+ & 164 & pass@1 on extended tests & 1,024 \\
LiveCodeBench V5+V6 & 342 (167+175) & Official fast/lite code-generation pass@1 & 2,000 \\
LongBench & 4,750 & Equal mean of 21 native subtask scores & 32/64/128/512 \\
MATH500 & 500 & Mathematical answer equivalence using Math-Verify & 8,192 \\
GPQA-Diamond & 198 & Choice exact match with flexible extraction after zero-shot reasoning & 8,192 \\
\bottomrule
\end{tabularx}
\par\smallskip
{\raggedright\footnotesize\noindent\emph{Note.} Question counts refer to the full selected subsets. Output limits are maximum generated-token counts and vary by subtask for LongBench.\par}
\end{table*}

The seven reported subsets cover mathematical reasoning, instruction compliance, code generation, long-context understanding, and expert knowledge. We evaluate the 342 problems added in the \href{https://github.com/LiveCodeBench/LiveCodeBench}{official LiveCodeBench releases} \texttt{release\_v5} (167) and \texttt{release\_v6} (175). We evaluate on all 21 datasets in the original LongBench benchmark. LongBench uses a total budget of 32,768 tokens, with middle truncation for overlong prompts. The overall score is the unweighted mean of the seven benchmark scores on a 0–100 scale.

\subsection{Baselines and precision settings}
\label{sec:4.2}

We use unquantized BF16 as the reference and compare nominal 1-bit configurations of KIVI \citep{kivi}, OSCAR \citep{oscar}, VecInfer \citep{vecinfer}, and TaSQ \citep{tasq}. KIVI retains its published group size and residual length, and OSCAR retains its published sink and recent-window sizes.

We use d8b8 and d8b10 to denote eight-dimensional vectors with 8-bit and 10-bit codeword indices, respectively. These correspond to 1.00 and 1.25 index bits per dimension. KIVI reduces its asymmetric codes from 2 bits to 1 bit. OSCAR replaces its INT2 payload with 1-bit scalar codes and uses eight-token refresh batches. VecInfer uses d8b8 for both keys and values, with a 16-token prefix and a dynamic recent buffer of 112 tokens. For all three models, we adapt the implementation accompanying the TaSQ paper to d8b8, with 256 codewords per eight-channel group for K and V, a 64-token recent window, and refreshes in eight-token blocks.

CALFA-KV uses the same recent-window and refresh settings as TaSQ. CALFA-KV (d8b8) and the adapted TaSQ baseline both use a nominal index rate of 1.00 bit per channel. CALFA-KV uses joint calibration with no fixed prefix and a budget of 56 additional FP16 anchors per layer and KV head. The nominal rates exclude scale metadata, codebooks, windows, and anchors.

\subsection{Benchmark results}
\label{sec:4.3}

Across all three models in Tables~\ref{tab:4}--\ref{tab:6}, CALFA-KV (d8b10) comes within 2.01 points of BF16 on the seven-benchmark mean at a nominal index rate of 1.25 bits per channel. The mean scores are 66.05 versus 66.95 on Qwen3-4B-Instruct-2507, 48.96 versus 50.26 on Llama-3.1-8B-Instruct, and 69.10 versus 71.11 on Qwen3-30B-A3B-Instruct-2507, corresponding to gaps of 0.90, 1.30, and 2.01 points, respectively. With the same budget of up to 56 additional FP16 anchors per layer and KV head, d8b10 improves on d8b8 by 1.84--2.38 points across the three models.

\begin{table*}[t]
\centering
\small
\setlength{\tabcolsep}{3pt}
\caption{Benchmark scores for Qwen3-4B-Instruct-2507}
\label{tab:4}

\resizebox{\textwidth}{!}{%
\begin{tabular}{l*{9}{r}}
\toprule
Method
& \shortstack{BPE}
& GSM8K
& IFEval
& HumanEval+
& \shortstack{LiveCodeBench\\V5+V6}
& LongBench
& MATH500
& \shortstack{GPQA-\\Diamond}
& \shortstack{Mean\\score} \\
\midrule
BF16
& 16.000 & 92.34 & 81.15 & 69.51 & 28.36 & 42.57 & 92.60 & 62.12 & 66.95 \\
VecInfer
& 1.000 & 1.06 & 34.20 & 21.95 & 0.00 & 24.05 & 1.60 & 13.64 & 13.79 \\
KIVI
& 2.000 & 0.45 & 47.87 & 17.07 & 0.00 & 8.43 & 3.40 & 9.09 & 12.33 \\
OSCAR
& 1.125 & 71.19 & 77.82 & 65.85 & 5.26 & 21.13 & 27.20 & 14.65 & 40.44 \\
TaSQ
& 1.008 & 88.48 & 70.98 & 63.41 & 18.71 & 39.87 & 83.20 & 52.53 & 59.60 \\
\rowcolor{blue!10}
\textbf{CALFA-KV (d8b10)}
& 1.258 & 89.76 & 83.73 & 69.51 & 27.77 & 41.72 & 90.80 & 59.09 & 66.05 \\
\rowcolor{blue!10}
\textbf{CALFA-KV (d8b8)}
& 1.008 & 89.84 & 82.99 & 67.07 & 22.80 & 41.15 & 89.80 & 54.54 & 64.03 \\
\bottomrule
\end{tabular}%
}

\par\smallskip
{\raggedright\footnotesize\noindent\emph{Note.} Mean is the unweighted average of the seven benchmark scores. Base BPE counts K/V payload and quantization metadata in bits per K/V element, excluding codebooks, prefix and recent windows, and anchors.\par}
\end{table*}

\begin{table*}[t]
\centering
\small
\setlength{\tabcolsep}{3pt}
\caption{Benchmark scores for Llama-3.1-8B-Instruct}
\label{tab:5}

\resizebox{\textwidth}{!}{%
\begin{tabular}{l*{8}{r}}
\toprule
Method
& GSM8K
& IFEval
& HumanEval+
& \shortstack{LiveCodeBench\\V5+V6}
& LongBench
& MATH500
& \shortstack{GPQA-\\Diamond}
& \shortstack{Mean\\score} \\
\midrule
BF16
& 79.76 & 74.68 & 57.93 & 14.33 & 46.88 & 47.40 & 30.81 & 50.26 \\
VecInfer
& 11.14 & 42.51 & 27.44 & 2.92 & 36.20 & 11.80 & 15.15 & 21.02 \\
KIVI
& 2.05 & 40.30 & 18.29 & 0.00 & 10.53 & 1.60 & 11.62 & 12.06 \\
OSCAR
& 76.80 & 68.95 & 56.71 & 7.60 & 21.32 & 37.60 & 13.13 & 40.30 \\
TaSQ
& 73.01 & 50.65 & 46.34 & 7.89 & 43.55 & 39.20 & 22.22 & 40.41 \\
\rowcolor{blue!10}
\textbf{CALFA-KV (d8b10)}
& 79.45 & 77.08 & 54.88 & 14.33 & 46.01 & 45.20 & 25.76 & 48.96 \\
\rowcolor{blue!10}
\textbf{CALFA-KV (d8b8)}
& 76.42 & 74.68 & 54.88 & 11.99 & 45.00 & 41.60 & 25.25 & 47.12 \\
\bottomrule
\end{tabular}%
}

\par\smallskip
{\raggedright\footnotesize\noindent\emph{Note.} Mean is the unweighted average of the seven benchmark scores.\par}
\end{table*}

\begin{table*}[t]
\centering
\small
\setlength{\tabcolsep}{3pt}
\caption{Benchmark scores for Qwen3-30B-A3B-Instruct-2507}
\label{tab:6}

\resizebox{\textwidth}{!}{%
\begin{tabular}{l*{8}{r}}
\toprule
Method
& GSM8K
& IFEval
& HumanEval+
& \shortstack{LiveCodeBench\\V5+V6}
& LongBench
& MATH500
& \shortstack{GPQA-\\Diamond}
& \shortstack{Mean\\score} \\
\midrule
BF16
& 95.83 & 82.81 & 71.34 & 35.67 & 47.05 & 96.40 & 68.69 & 71.11 \\
VecInfer
& 2.50 & 35.30 & 31.10 & 1.17 & 28.84 & 1.80 & 15.66 & 16.62 \\
KIVI
& 0.08 & 47.32 & 20.12 & 0.00 & 9.14 & 2.40 & 14.65 & 13.39 \\
OSCAR
& 59.74 & 74.68 & 68.29 & 6.43 & 22.44 & 24.40 & 11.62 & 38.23 \\
TaSQ
& 93.18 & 69.69 & 65.24 & 22.22 & 43.33 & 87.40 & 55.05 & 62.30 \\
\rowcolor{blue!10}
\textbf{CALFA-KV (d8b10)}
& 95.45 & 81.52 & 68.90 & 33.92 & 46.46 & 93.80 & 63.64 & 69.10 \\
\rowcolor{blue!10}
\textbf{CALFA-KV (d8b8)}
& 94.92 & 80.78 & 68.29 & 28.36 & 44.70 & 91.40 & 58.59 & 66.72 \\
\bottomrule
\end{tabular}%
}

\par\smallskip
{\raggedright\footnotesize\noindent\emph{Note.} Mean is the unweighted average of the seven benchmark scores.\par}
\end{table*}

At a nominal index rate of 1.00 bit per channel, CALFA-KV (d8b8) improves on the adapted TaSQ baseline by 4.43, 6.71, and 4.42 points in seven-benchmark mean score, in the same model order. It scores higher than TaSQ on all 21 model--benchmark pairs, with the largest gains on IFEval (11.09--24.03 points).

For Qwen3-4B-Instruct-2507, CALFA-KV shares one FP16 key scale across eight KV heads per token and layer, adding $16/[8(128+128)] \approx 0.008$ BPE to the nominal index rates. Table~\ref{tab:4} reports a base BPE of 1.008 for d8b8, matching the adapted TaSQ baseline, and an estimated 1.258 for d8b10, assuming packed 10-bit indices and the same scale layout.

\subsection{Layerwise error analysis}
\label{sec:layerwise-error-propagation}

\begin{figure*}[t]
    \centering
    \includegraphics[width=\textwidth]{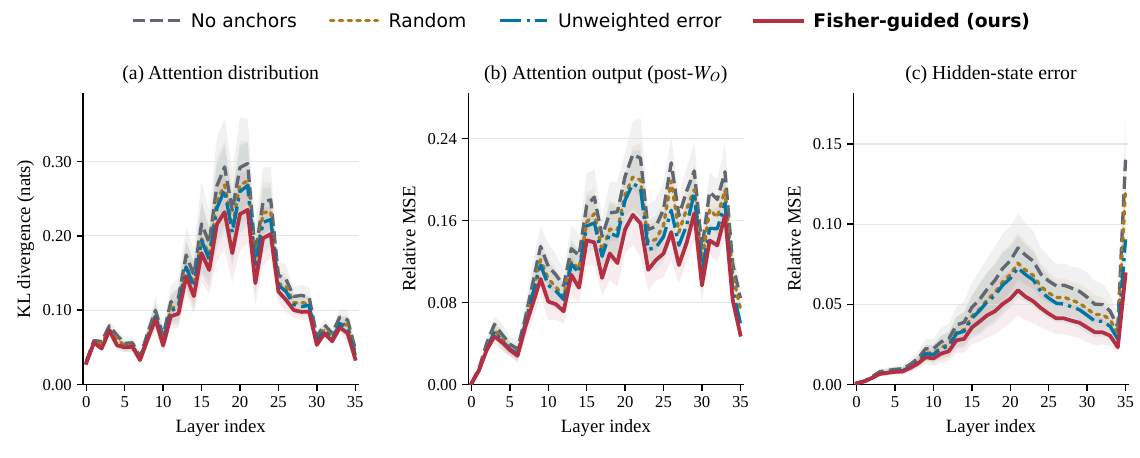}
    \caption{Layerwise errors relative to a BF16 reference on Qwen3-4B-Instruct-2507 with a fixed, jointly calibrated d8b8 quantizer. Panels show (a) attention-distribution KL divergence (nats), (b) relative MSE of the attention output after $W_O$, and (c) hidden-state relative MSE. Relative MSE is the sum of squared differences divided by the reference signal energy. No anchors uses zero additional anchors; Random, Unweighted error, and Fisher-guided (ours) each use a budget of 56 additional K/V pairs per layer and KV head. Curves average 24 prompts, with Random averaged over three seeds. Shaded bands show 95\% prompt-bootstrap confidence intervals. Lower is better.}
    \label{fig:layerwise_error_propagation}
\end{figure*}

We freeze the jointly calibrated d8b8 quantizer and vary only the anchor policy in Figure~\ref{fig:layerwise_error_propagation}. No anchors is the quantized control. Random selects K/V pairs at random, Unweighted error ranks pairs by squared K/V reconstruction error, and Fisher-guided (ours) weights those errors by Fisher-type sensitivities estimated offline. The diagnostic uses eight prompts each from Qasper, GovReport, and LCC. For each 1,024-token prompt, all policies replay the same 128-token BF16 reference continuation using dense FP32 attention and BF16 model weights.

Across prompts and layers, Fisher-guided (ours) reduces attention-distribution KL, post-$W_O$ relative MSE, and hidden-state relative MSE by 18.4\%, 25.0\%, and 32.2\%, respectively, relative to No anchors. It also yields lower mean errors than Random and Unweighted error on all three metrics at the same anchor budget. The attention curves separate more around their mid-layer peaks than in the early layers. At the final layer, Fisher-guided selection brings hidden-state relative MSE to about half that of No anchors. These comparisons quantify intermediate distortion under fixed-history replay.

\subsection{Ablation of anchor selection}
\label{sec:4.4}

Table 5 compares anchor policies for Qwen3-4B-Instruct-2507. The Random and Fisher-guided variants share the same anchor budget, isolating the effect of the selection policy. JC + Fisher-guided corresponds to CALFA-KV (d8b8) in Table~\ref{tab:4}.

\begin{table*}[!t]
\centering\small
\setlength{\tabcolsep}{3pt}
\caption{Anchor selection ablation on Qwen3-4B-Instruct-2507}
\label{tab:7}
\begin{tabular*}{\textwidth}{@{\extracolsep{\fill}}l*{8}{r}@{}}
\toprule
Configuration & GSM8K & IFEval & HumanEval+ & \shortstack{LiveCodeBench\\V5+V6} & LongBench & MATH500 & \shortstack{GPQA-\\Diamond} & \shortstack{Mean\\score} \\
\midrule
BF16 & 92.34 & 81.15 & 69.51 & 28.36 & 42.57 & 92.60 & 62.12 & 66.95 \\
JC & 88.78 & 76.15 & 65.85 & 21.05 & 41.19 & 87.80 & 57.07 & 62.56 \\
JC + Random & 88.93 & 78.92 & 66.46 & 22.51 & 30.00 & 88.20 & 58.58 & 61.94 \\
JC + Fisher-guided & 89.84 & 82.99 & 67.07 & 22.80 & 41.15 & 89.80 & 54.54 & 64.03 \\
\bottomrule
\end{tabular*}
\par\smallskip
{\raggedright\footnotesize\noindent\emph{Note.} JC denotes joint calibration without additional anchors. All quantized variants share a jointly calibrated d8b8 quantizer and a 64-token recent window. Each anchor policy has a budget of 56 additional FP16 K/V pairs per layer and KV head. Mean is the unweighted average of the seven benchmark scores.\par}
\end{table*}

Relative to JC alone, random retention lowers the mean score by 0.62 points, whereas score-based retention raises it by 1.47 points. At the same anchor count, score-based selection exceeds random selection by 2.09 points and scores higher on six of seven tasks, including LongBench (41.15 versus 30.00). This comparison supports the proposed selection rule over random selection for this configuration.

\section{Conclusion}
\label{sec:5}

CALFA-KV combines joint quantizer calibration with Fisher-guided anchor selection. Offline, it jointly calibrates the pre-RoPE key center, radial scale parameters, and K/V codebooks against a shared objective on each layer’s projected attention output. During inference, it weights the calibrated quantizer’s squared K/V reconstruction errors by offline-estimated Fisher-type sensitivities and retains the highest-scoring KV states in FP16. 

At a nominal index rate of 1.25 bits per channel and with up to 56 additional FP16 anchors per layer and KV head, CALFA-KV (d8b10) achieves mean scores across seven benchmarks only 0.90–2.01 points below the BF16 KV-cache baseline on the three evaluated models.

\clearpage
\bibliographystyle{CALFAmla}
\bibliography{references}
\clearpage
\typeout{CALFA-MAIN-PAGES: \number\numexpr\value{page}-1\relax}
\onecolumn
\appendices
\section{Anchor error and selection guarantees}
\label{sec:A}

\subsection{Restoration error and loss expansion}
\label{sec:A.1}

Fix one layer and one KV head with $m$ eligible historical KV states. For each KV state, hold its stored scale and the finalized $\mu,D,P$ fixed. Stack its normalized key coordinates from Eq.~\eqref{eq:1} and original value coordinates in token, side, and channel order as $\xi_x\in\mathbb R^{m(d_K+d_V)}$. The loss $\ell_x$ below is expressed in these scoring coordinates through their fixed affine inverse. Exact prefix/recent/live KV states are outside this vector.

Hold the reference states, quantizer, stored scales, and eligible set fixed while comparing anchor sets. Let $Q_\theta(\xi_x)$ denote actual payload reconstruction expressed back in the same scoring coordinates, including the inverse positional transform in Section~\ref{sec:3.3}. Its error is $\delta_x=Q_\theta(\xi_x)-\xi_x$. Let $P_B$ be the diagonal orthogonal projector onto the K/V coordinates of selected KV states $B$. The error after restoring the selected KV states is

\[
e_{x,B}=(I-P_B)\delta_x+r_{x,B},\qquad P_Br_{x,B}=r_{x,B}.
\tag{A1}
\label{eq:A1}
\]

For the main construction, the saved KV state is restored at reference precision and $r_{x,B}=0$. Holding the rest of the model fixed, the expected change in request loss under an online policy $\pi$ is

\[
J_{\mathrm{true}}(\pi)=\mathbb E_X\!\left[\ell_X(\xi_X+e_{X,\pi(X)})-\ell_X(\xi_X)\right].
\tag{A2}
\label{eq:A2}
\]

The online policy uses only the K/V states and scores available at admission. For a fixed request, assume $\ell$ is twice continuously differentiable near the segment from $\xi$ to $\xi+e$, and let $g=\nabla\ell(\xi)$ and $H=\nabla^2\ell(\xi)$. The second-order expansion is

\[
\ell(\xi+e)-\ell(\xi)=g^{\mathsf T}e+\tfrac12e^{\mathsf T}He+R_3(e),
\tag{A3}
\label{eq:A3}
\]

where $R_3(e)$ is the Taylor remainder. Online scoring instead uses the conditional mean square of $g^{\mathsf T}e$ in Equation~\eqref{eq:A4}, approximated by pooled offline squared-gradient weights and realized reconstruction errors. The bound in Section~\ref{sec:A.3} concerns this scoring criterion, rather than the true loss change in Equation~\eqref{eq:A3} or the expected loss in Equation~\eqref{eq:A2}.

\subsection{Gradient moments and conditional risk}
\label{sec:A.2}

Let $g=\nabla_\xi\ell$ be the reference activation gradient with finite second moment, and let $\mathcal I$ be the information available to the online selector, including the eligible KV states, their reconstruction errors, and the fixed assets. For an $\mathcal I$-measurable reconstruction error $e$, the conditional squared linear perturbation is

\[
F_{\mathcal I}=\mathbb E[gg^{\mathsf T}\mid\mathcal I],\qquad
\mathcal R_{\mathcal I}(e)=\mathbb E[(g^{\mathsf T}e)^2\mid\mathcal I]=e^{\mathsf T}F_{\mathcal I}e.
\tag{A4}
\label{eq:A4}
\]

Here $F_{\mathcal I}$ is the full gradient second moment conditioned on the current candidate information, in the same coordinates as $e$. The deployed score in Equation~\eqref{eq:8} uses pooled and regularized channel weights $\overline F$ instead (Section~\ref{sec:3.3} and Appendix~\ref{sec:B.2}).

For a positive conditional model distribution $p(y\mid x,\xi)$ at a fixed working point, assume parameter-independent support, twice differentiability, and valid interchange of differentiation and integration. The model Fisher identity is

\[
\mathbb E_{Y\sim p}[ss^{\mathsf T}]=-\mathbb E_{Y\sim p}[\nabla_\xi^2\log p(Y)],\qquad
s=\nabla_\xi\log p(Y).
\tag{A5}
\label{eq:A5}
\]

Calibration instead uses observed labels and request-specific working points. Its empirical gradient second moments provide Fisher-type weights for Equation~\eqref{eq:A4}; the identity above does not equate these deployed estimates with the Hessian of a particular request's loss.

\subsection{Additive-score approximation bound}
\label{sec:A.3}

With exact restoration and a fixed eligible set $\mathcal M$, the deployed diagonal proxy is the sum of scores of unprotected KV states:

\[
\widehat{\mathcal R}(B)=\sum_{t\in\mathcal M\setminus B}S_t,\qquad |B|\le N.
\tag{A6}
\label{eq:A6}
\]

The equal-cost, nonnegative-score case is minimized by the highest-scoring $\min(N,|\mathcal M|)$ KV states, as proved in Section A.4. Let $F_{\mathrm{off}}$ be a same-dimensional offline reference gradient second moment for this fixed candidate layout, and let $D_{\mathrm{off}}=\operatorname{diag}(F_{\mathrm{off}})$, where $\operatorname{diag}$ retains the diagonal as a matrix. Repeating the deployed channel weights across token coordinates gives $\overline F_i$, the coordinate-level form of the $\overline F^K$ and $\overline F^V$ weights in Equation~\eqref{eq:8}. Thus the analysis compares $F_{\mathcal I}$ with $F_{\mathrm{off}}$, its diagonal $D_{\mathrm{off}}$, and finally the deployed diagonal weights $\overline F_i$. The full reference matrix is used for analysis; a different candidate dimension uses the corresponding matrix. The error budget below accounts for these three steps:

\[
\begin{aligned}
\varepsilon_x={}&\left(\|F_{\mathcal I}-F_{\mathrm{off}}\|_{\mathrm{op}}
+\|F_{\mathrm{off}}-D_{\mathrm{off}}\|_{\mathrm{op}}\right)\|\delta_x\|_2^2\\
&+\sum_i|(F_{\mathrm{off}})_{ii}-\overline F_i|\delta_{x,i}^2.
\end{aligned}
\tag{A7}
\label{eq:A7}
\]

If $\widehat B$ minimizes the proxy and $B^\star$ minimizes conditional risk over the same eligible sets, the uniform bound gives

\[
\mathcal R_{\mathcal I}(e_{x,\widehat B})\le\mathcal R_{\mathcal I}(e_{x,B^\star})+2\varepsilon_x.
\tag{A8}
\label{eq:A8}
\]

Both optimizers range over $\{B\subseteq\mathcal M:|B|\le N\}$, with the same frozen reference states. The three terms in $\varepsilon_x$ account for moment shift, omitted correlations, and deployed-weight differences. We do not estimate them or establish that this bound is small for the reported models.

\textbf{Proof.} Write $F=F_{\mathcal I}$, $F_0=F_{\mathrm{off}}$, $D_0=\operatorname{diag}(F_0)$, and $\overline D=\operatorname{diag}(\overline F)$. Inserting $F_0$ and $D_0$ and using $|e^{\mathsf T}Me|\le\|M\|_{\mathrm{op}}\|e\|_2^2$ gives

\[
\begin{aligned}
|e^{\mathsf T}Fe-e^{\mathsf T}\overline D e|\le{}&\|F-F_0\|_{\mathrm{op}}\|e\|_2^2
+\|F_0-D_0\|_{\mathrm{op}}\|e\|_2^2\\
&+\sum_i|(F_0)_{ii}-\overline F_i|e_i^2.
\end{aligned}
\tag{A9}
\label{eq:A9}
\]

In the main exact-restoration construction, $e_B=(I-P_B)\delta$. Coordinate deletion gives $\|e_B\|_2\le\|\delta\|_2$ and $e_{B,i}^2\le\delta_i^2$, hence Equation~\eqref{eq:A7} bounds Equation~\eqref{eq:A9} uniformly for every feasible $B$. For a proxy minimizer $\widehat B$ and a conditional-risk minimizer $B^\star$ on the same finite feasible family,

\[
\mathcal R_{\mathcal I}(e_{\widehat B})\le\widehat{\mathcal R}(\widehat B)+\varepsilon_x
\le\widehat{\mathcal R}(B^\star)+\varepsilon_x
\le\mathcal R_{\mathcal I}(e_{B^\star})+2\varepsilon_x.
\tag{A10}
\label{eq:A10}
\]

This proves Equation~\eqref{eq:A8} for the conditional mean-squared linear risk in Equation~\eqref{eq:A4}. The argument also applies to a vector containing multiple heads, using the product of their budget constraints and including cross-head correlations in the full moment matrices.

\subsection{Top-budget and streaming selection}
\label{sec:A.4}

For a KV state restored with residuals $r^K_t,r^V_t$, the diagonal proxy contribution after restoration is $T_t$ and its net gain is $G_t$:

\[
T_t=\sum_d\overline F^K_d(r^K_{t,d})^2+\sum_d\overline F^V_d(r^V_{t,d})^2,\qquad G_t=S_t-T_t.
\tag{A11}
\label{eq:A11}
\]

In the main construction, exact restoration gives $T_t=0$ and $G_t=S_t$. For a format with restoration error confined to each KV state, select KV states by positive net gain under an upper-bound budget.

For the top-budget selection rule, minimizing the remaining sum equals maximizing the sum removed. With nonnegative scores, a feasible set can be extended to size $\min(N,|\mathcal M|)$ without worsening the objective. A selected lower-score element can then be exchanged with an unselected higher-score element. Repeating exchanges reaches a top-budget set. Ties permit multiple optima; a position-based order determines a reproducible representative. If $N=0$, the only feasible set is empty.

\textbf{Streaming invariant.} Let $V$ be the previously admitted history and assume $B_{\mathrm{old}}=\operatorname{TopN}(V)$ under a fixed total order. When $|V|<N$, every KV state is retained. Otherwise, all $N$ KV states of $B_{\mathrm{old}}$ precede any omitted KV state and remain in $B_{\mathrm{old}}\cup U$. An omitted KV state therefore cannot enter the top-$N$ of $V\cup U$. Equation~\eqref{eq:9} therefore returns that top-$N$ set. Induction extends the result across admissions with fixed scores and a common score definition.

With restoration errors confined to individual KV states, the exchange argument selects the largest positive $G_t$ under an upper-bound budget. An exactly filled budget selects the largest feasible $N$ net benefits.

\section{Coordinates, weights, and attention replacement}
\label{sec:B}

\subsection{Transforming errors and sensitivity}
\label{sec:B.1}

We use the normalized key coordinates of Section~\ref{sec:3.3} and the original value coordinates for scoring. Suppress the layer and head indices. Holding the center, channel weights, permutation, and each token's stored scale fixed gives the affine key map $z_t=J_t(k_t^{\mathrm{pre}}-\mu)$, where

\[
\begin{aligned}
J_t&=\frac{PD}{\operatorname{FP32}(n_t)},\\
e_t^z&=J_t e_t^{\mathrm{pre}},\qquad
g_t^z=J_t^{-\mathsf T}g_t^{\mathrm{pre}},\\
F_t^z&=J_t^{-\mathsf T}F_t^{\mathrm{pre}}J_t^{-1},\\
(e_t^z)^{\mathsf T}F_t^z e_t^z
&=(e_t^{\mathrm{pre}})^{\mathsf T}F_t^{\mathrm{pre}}e_t^{\mathrm{pre}}.
\end{aligned}
\tag{A12}
\label{eq:A12}
\]

Here $F_t^{\mathrm{pre}}$ and $F_t^z$ are matching full gradient second moments at a fixed working point. The chain rule gives $g_t^z=\operatorname{FP32}(n_t)PD^{-1}g_t^{\mathrm{pre}}$. This is a fixed-scale coordinate transformation, rather than the full Jacobian of the input-dependent map in Eq.~\eqref{eq:1}. We transform signed gradients before squaring and pooling, using the stored scale of each calibration token. Value gradients and errors require no coordinate transform.

Actual key errors are measured after native reconstruction. We invert the positional transform using the cached sine and cosine values, including their squared-norm correction, then apply $J_t$ to the pre-RoPE error. The center cancels in this difference. Changing the sign of both error definitions leaves the squared score unchanged. The fixed-coordinate construction does not replace the native decoder used to measure the errors.

Equation~\eqref{eq:A12} preserves the full quadratic form when errors and second moments are transformed together. A fixed diagonal weight pooled in one coordinate system cannot generally be reused in another: $n_t$ varies across tokens, and a post-RoPE change of basis also mixes channels. The deployed estimates must therefore identify their differentiation tensors, fixed-scale transform, and codec assets. In Appendix A, $\xi_x$ stacks these scoring coordinates and the reference loss is composed with their fixed inverse coordinate map.

\subsection{Regularized sensitivity weights}
\label{sec:B.2}

For a reference sequence $y_i$ of length $L_i$, the accompanying implementation uses all shifted next-token targets, $\mathcal T_i=\{2,\ldots,L_i\}$, in the mean loss $\ell_i=-|\mathcal T_i|^{-1}\sum_{j\in\mathcal T_i}\log p_{\mathrm{ref}}(y_{i,j}\mid y_{i,<j})$. It captures K after model-native key normalization and before RoPE, and V at the value-projection output. All captured positions $\mathcal M_i=\{1,\ldots,L_i\}$ enter the pooling count, including positions with zero gradient. The loss-position set and captured-state set are distinct from the online anchor-eligible history.

We differentiate this mean loss with respect to the reference states, cast the signed gradients to FP32, apply the fixed-scale key transformation in Section B.1, and only then square and pool as in Eq.~\eqref{eq:7}. Thus each captured position contributes once, while changing the loss positions or loss reduction changes the estimates. The supervised sequences, masks, and any overlap with Fit/Select inputs must be recorded. These are calibration inputs; the sensitivity loss is separate from Eq.~\eqref{eq:6}.

For each layer and head, begin with nonnegative raw channel estimates $f^K_d,f^V_d$. With a quantile level $q\in(0,1]$, clip each side along its channel dimension to its own upper quantile. Let $\widetilde f^c$ be these clipped estimates, $m$ the average of the two channel means, and $\rho\in[0,1]$ the shrinkage coefficient. The reference rule is

\[
\begin{aligned}
f^{c,\mathrm{sh}}_d&=(1-\rho)\widetilde f^c_d+\rho m,\\
\overline F^c_d&=\frac{f^{c,\mathrm{sh}}_d}{\tfrac12(\operatorname{mean}_d f^{K,\mathrm{sh}}_d+\operatorname{mean}_d f^{V,\mathrm{sh}}_d)}.
\end{aligned}
\tag{A13}
\label{eq:A13}
\]

The denominator must be finite and positive. The accompanying implementation uses clipping quantile $q=0.995$ and shrinkage $\rho=0.05$. Clipping and shrinkage can change token rankings; a final common positive scale within one layer/head preserves its ranking. Because each head has its own normalization, the selection guarantee applies within each head's budget.

\subsection{Exact KV state restoration in decode}
\label{sec:B.3}

For one post-RoPE query $q\in\mathbb R^{d_K}$, let $d=d_K$. All sums and anchor replacements below range over positions visible to this query. Attention forms logits from post-RoPE keys, applies softmax, and aggregates values:

\[
a_t=\frac{q^{\mathsf T}k^{\mathrm{post}}_t}{\sqrt d},\qquad
p=\operatorname{softmax}(a),\qquad o=\sum_t p_tv_t.
\tag{A14}
\label{eq:A15}
\]

The projected output also applies $W_O$ to the concatenated head outputs. Restoring an anchor changes both its logit and the value entering this aggregation.

Let a baseline reconstructed cache yield logits $\widehat a_t$, values $\widehat v_t$, normalizer $Z_0=\sum_t\exp(\widehat a_t)$, and output $o_0$. For selected anchors, let $a^\star_t,v^\star_t$ be the restored logits and values. In exact arithmetic, restoring these KV states gives

\[
\begin{aligned}
Z_\star&=Z_0+\sum_{t\in B}[\exp(a^\star_t)-\exp(\widehat a_t)],\\
Y_\star&=Z_0o_0+\sum_{t\in B}[\exp(a^\star_t)v^\star_t-\exp(\widehat a_t)\widehat v_t],\\
o_\star&=Y_\star/Z_\star.
\end{aligned}
\tag{A15}
\label{eq:A16}
\]

A common maximum-logit shift controls the exponential range but does not prevent cancellation when a dominant reconstructed anchor contribution is subtracted from a rounded total. A stable reference path instead excludes selected anchors from the reconstructed partition, then combines their positive exponential-weight sums and corresponding weighted-value sums using a common shift or log-sum-exp. The value sums may have signed components. A fused implementation must be checked against this reference; Equation~\eqref{eq:A16} alone is not a numerically stable update recipe. The identity gives attention over the hybrid cache with selected KV states restored. Its implementation uses the causal mask, grouped-query head mapping, and matching original/reconstructed coordinates. During chunked prefill, the read snapshot from Appendix C instead replaces reconstructed historical cache entries before attention, alongside exact recent/live cache entries.

\section{Calibration and implementation details}
\label{sec:C}

\subsection{Coordinate construction and calibration updates}
\label{sec:C.1}

The representation in Section~\ref{sec:3.1} is parameterized by head dimensions and codebook size. The affine codec and anchor interfaces described below are specialized to the d8b8 configuration of Qwen3-4B-Instruct-2507, with 36 layers, eight KV heads, $d_K=d_V=128$, $g=8$, and $K_c=256$.

\paragraph{Deterministic coordinate construction}
The scale reduction consumes $x=\operatorname{FP32}(A(k-\mu))$ in head-major, pre-permutation channel order. Writing $\operatorname{fl}_{32}$ for one FP32 operation, its balanced tree is

\[
a_i^{(0)}=\operatorname{fl}_{32}(x_i^2),\qquad
a_i^{(r+1)}=\operatorname{fl}_{32}
\!\left(a_{2i}^{(r)}+a_{2i+1}^{(r)}\right).
\]

An unmatched entry is carried unchanged to the next level. After the final level, division by $D_K$, square root, and the lower bound are evaluated in FP32, followed by one FP16 cast. All later consumers promote this stored value; none substitutes an unrounded norm. At coordinate initialization, $A=D$ and $\mu=0$. Coordinate statistics, codebook fitting, calibration replay, cache writes, and inference all consume this same stored scale. Since $P$ is applied only after the scale is formed, changing the grouping cannot change $n_t$.

For $P$, each RoPE-coupled coordinate unit is indivisible. Hierarchical minimum-cost perfect matching under Eq.~\eqref{eq:5} merges units to width $g$. The source uses $\epsilon_{\mathrm{abs}}=10^{-30}$, $\epsilon_{\mathrm{rel}}=10^{-6}$, $\epsilon=10^{-4}$, and $\lambda_h=\max(10^{-30},10^{-4}\operatorname{tr}(\Sigma_h)/d_K)$. The matching cost initializes a grouping; it does not optimize the final finite-codebook distortion.

\paragraph{Finite-rate regrouping and pair swaps}
The initialization provenance records reconstruction-based regrouping followed by pair-swap refinement. The first stage evaluates two passes of legal regroupings with two fixed random seeds. Inner selection uses a bounded codebook-fitting proxy; outer selection refits both the original and proposed groups with the full fitting budget. For each seed, let $r_E$ and $r_F$ be the proposed-to-refitted-original ratios of ordinary and Fisher-weighted physical-coordinate reconstruction SSE. The mean ratio $(r_E+r_F)/2$ must be at most $0.99$ for inner selection and $0.9975$ for outer selection, while each individual ratio must be at most $1.005$. The outer check also requires fixed-V attention-context SSE no greater than $1.005$ times the refitted-original value. These attention checks use a local replay proxy.

The later pair-swap stage compares each proposal with both the current P/C result and a no-swap control that refits the affected codebooks. Proposal and control use the same calibration samples, initialization rule, random seeds, and maximum fitting budget; data-dependent k-means++ sampling need not select identical initial codewords. Acceptance requires attention-distribution KL no greater than $0.995$ times either comparator. For each comparator, GPQA-stratum KL, code-stratum KL, and fixed-V attention-output relative SSE may be at most $1.01$ times its value plus $10^{-12}$. Physical-coordinate MSE and Fisher-MSE are diagnostic quantities in this stage, rather than acceptance requirements. The exact accepted permutation and codebooks are retained together without a post-selection refit. The final $D/P$ remain fixed during continuous calibration.

\paragraph{Native reconstruction precision}
Equation~\eqref{eq:3} specifies a real-arithmetic coordinate relationship. The native decode path unweights and rotates the selected key codewords, rounds the key dot operands to FP16, accumulates the query dot products, and then multiplies by the stored FP16 scale promoted to FP32. The rotated center dot product is computed separately and added without this scale factor. The candidate decoder mirrors these operand-rounding and center-restoration steps for scoring. We invert the model's cached positional transform when expressing that actual reconstruction error in normalized coordinates. Algebraically equivalent operation orders need not give identical floating-point errors, so hard calibration acceptance and anchor scoring use the deployed reconstruction path.

\paragraph{Calibration windows and evaluated queries}
The accompanying implementation uses 64 fixed windows in both codebook warm-start and final joint calibration. With zero-based window indices $w=0,\ldots,63$, Fit contains the 48 windows with $w\bmod4\ne3$, and Select contains the remaining 16. The codebook-only warm-start uses 2048-token windows and query positions $127+128j$, $j=0,\ldots,15$; its historical region for query position $s$ is $t<s+1-64$. Final joint calibration uses the template-wrapped 2056-token windows and query positions $135+128j$ for the same 16 values of $j$. Its historical lengths come from the captured per-query cache metadata, including the native page-8 flush boundaries. Anchors are disabled during joint calibration. In Eq.~\eqref{eq:6}, each output vector concatenates only the evaluated query positions and all projected channels in a fixed order; the SSE and reference energy exclude outputs at other query positions.

The radial weights remain positive when each RoPE pair shares one log-coordinate $u^A$. Writing $\pi(j)$ for the pair containing channel $j$, we parameterize the weights and center as

\[
a_{h,j}=d_{h,j}\exp\{\gamma(u^A_{h,\pi(j)}-\overline u^A)\},\qquad \mu_i=s_i u^\mu_i.
\tag{A16}
\label{eq:A17}
\]

Here $A_h$ and $D_h$ are the head-$h$ blocks of $A$ and $D$, with $a_{h,j}=[A_h]_{jj}$ and $d_{h,j}=[D_h]_{jj}=w_{h,j}$. The quantity $s_i$ is a fixed, floored RMS from fitting data, $\gamma$ sets the radial coordinate unit, and $\overline u^A$ is the layer mean of pair coordinates. Codebook updates use fixed initial per-group RMS units. Collect the normalized center, radial, and codebook coordinates in $u$, and write the physical parameters as $\theta=\Phi(u)$. Center and radial parameters remain FP32; codebooks are converted to their declared storage dtype for candidate evaluation.

Nearest-codeword assignments and stored FP16 scales make the deployed objective discontinuous. At an accepted parameter point $u_0$, hold its assignments fixed and align the continuous scale with the stored scale $n_{t,0}$:

\[
\begin{aligned}
\widetilde n_t(u)&=n_{t,0}+\rho_t(u)-\rho_t(u_0),\\
\rho_t(u)&=\max\!\left(\epsilon,\frac{\|A(u)(k_t-\mu(u))\|_2}{\sqrt{D_K}}\right).
\end{aligned}
\tag{A17}
\label{eq:A18}
\]

For fixed assignments and finite $E_l>0$, concatenate the continuous replay output residuals over fitting windows and divide by $\sqrt{E_l}$ to define $\widetilde r(u)$. Its token scale is aligned by Equation~\eqref{eq:A18}; the entire floating-point replay is not asserted to equal the stored-precision computation. Away from scale-floor boundaries, let $J=\partial\widetilde r/\partial u$. At a boundary, use the implementation's declared autodiff convention to propose a step; native replay still decides acceptance. With $g_u=J^{\mathsf T}\widetilde r$, the damped Gauss–Newton proposal solves

\[
\begin{aligned}
(J^{\mathsf T}J+\lambda I)s&=-g_u,\\
\lambda&=\max\!\left(10^{-12},0.01\max\!\left(0,\frac{g_u^{\mathsf T}J^{\mathsf T}Jg_u}{g_u^{\mathsf T}g_u}\right)\right).
\end{aligned}
\tag{A18}
\label{eq:A19}
\]

A zero gradient ends the layer update before the ratio is evaluated. Jacobian-vector and vector-Jacobian products (JVP/VJP), together with conjugate gradients (CG), compute an approximate step $s$. The proposal rescales it to $d=\|u\|_2s/\|s\|_2$ and requires finite nonzero norms and $g_u^{\mathsf T}d<0$. The reported configuration uses at most eight layer-update rounds, at most 24 CG iterations with relative residual tolerance $10^{-4}$, and the backtracking sequence $\alpha_m=0.02\,2^{-m}$ for $m=0,\ldots,11$. Each trial is mapped through $\Phi$, cast to the deployment dtypes, and reassigned before native replay. Center and radial parameters remain FP32, while the K/V codebooks and key scale use FP16 storage. Fit and Select window identities are fixed before optimization. Acceptance is the conjunction of a valid finite replay, strictly lower Fit loss, and the configured non-degradation predicate on Select; those windows participate in model selection rather than independent held-out evaluation. The run configuration also records numerical floors, coordinate units, masks, and sensitivity windows.

\textbf{Algorithm 1 (offline joint calibration and sensitivity preparation).}

\begingroup\small
\begin{verbatim}
Input: frozen model, Fit/Select and sensitivity windows, parent codebooks,
       optimizer and acceptance configuration cfg
Output: stored codec, sensitivity weights Fbar
freeze(D, P, transforms, masks, dtypes, coordinate units)
for layer l = 1,...,L:
    ref <- capture_frozen(model, l, Fit, Select)
    u <- normalized_init(parent, mu=0, A=D)
    for round = 1,...,cfg.max_rounds:
        theta <- stored_parameters(Phi(u))
        ids, scale <- native_write(theta, ref)
        r <- aligned_residual(u, ids, scale, ref)
        d <- normalized_damped_GN_CG(r, u, cfg)
        if invalid_or_zero(d): break
        accepted <- false
        for alpha in cfg.backtracking_schedule:
            ucand <- u + alpha*d
            cand <- stored_parameters(Phi(ucand))
            out <- native_replay(cand, ref)
            if finite_valid(out) and improves_fit(out) and passes_checks(out, cfg):
                u <- ucand; accepted <- true
                break
        if not accepted: break
    codec[l] <- stored_parameters(Phi(u))
    sens <- capture_frozen(model, l, sensitivity_windows, cfg.loss_masks)
    gref <- signed_reference_gradients(sens)
    gscore <- fixed_scale_coordinate_transform(gref, codec[l], sens)
    Fbar[l] <- regularize(pool_squares(gscore))
check_coordinates(codec, Fbar)
\end{verbatim}
\endgroup

Here \texttt{native\_replay} refreshes encoding caches, recomputes stored FP16 scales and hard assignments, and evaluates projected outputs under the frozen masks. Sensitivity preparation follows Appendix~\ref{sec:B.1}--\ref{sec:B.2}; saved assets record the loss, captured positions, and coordinate identity.

\textbf{Algorithm 2 (online anchor admission and bounded retention).}

\begingroup\small
\begin{verbatim}
Input: request-fixed codec and Fbar, budgets N[l,h],
       native-encoded candidates U with request/layer/position/slot IDs
State: anchor directories B, initially empty for each request/layer/head
# U contains actual quantized prefill rows or native recent migrants.
# Native encoders have written U; original K/V remain readable.
for each request x and layer l with a candidate batch U[x,l]:
    Uvalid <- valid native candidate rows in U[x,l]
    for each row u in Uvalid:
        n[u] <- stored_scale(payload[u])
        rec[u] <- reconstruct_actual(payload[u], position[u], n[u], codec[l])
        for each KV head h:
            score[u,h] <- Eq(8)(row[u,h], rec[u,h], Fbar[l,h])
    for each KV head h:
        Bnew[h] <- TopN(B[h] union Uvalid, N[l,h])
        retain paired FP16 K/V from the old directory or candidate rows
        record the selected positions, quantized slots, and fixed scores
        B[h] <- Bnew[h]
    keep the quantized payload of every historical row
\end{verbatim}
\endgroup

In Algorithm 2, a cache row contains the K/V entries for one token position in one request and layer, across all KV heads. Admission begins after native encoding and retains original candidate K/V until scoring and selected-anchor copies finish. Eligibility follows the actual quantized slots and native flush plan; the prefill/extend and decode read/write order is specified in Section~\ref{sec:3.4}.

Key scales are fixed at the first cache-write boundary and retained with recent K/V; migration consumes them without recomputing an RMS. The policy uses a 64-token recent window with refreshes in eight-token blocks. Pending readers are synchronized before migration reuses buffers or changes read-visible mappings. Each attention operation observes stable cache contents and mappings, and updated anchor directories take effect for subsequent reads.

Selection uses a stable position-based tie rule; with restoration error, use positive net gain from Equation~\eqref{eq:A11}. The codec, coordinates, and weights remain fixed for each request. New assets require a new request and empty directories, since re-scoring retained states cannot recover the top-$N$ of all history after discarded originals are lost.

\subsection{Storage and scoring cost}
\label{sec:C.2}

For a layer with $n$ historical token positions, K/V dimensions $D_K,D_V$, index rates $b_K,b_V$, per-token scale metadata $m_s$, and per-head anchor budgets $N_h$, a storage inventory in bytes is

\[
\begin{aligned}
M_{\mathrm{layer}}\le{}&\frac{n(D_Kb_K+D_Vb_V)}8+n m_s\\
&+M_{\mathrm{window}}+M_{\mathrm{static}}+M_{\mathrm{alloc}}\\
&+\sum_h N_h\left(\frac{b_{\mathrm{hp}}(d_K+d_V)}8+m_{\mathrm{dir}}\right).
\end{aligned}
\tag{A19}
\label{eq:A20}
\]

Here $m_s$ and $m_{\mathrm{dir}}$ are bytes per token and per anchor entry, while $b_{\mathrm{hp}}$ is bits per stored high-precision channel. The terms account for the base quantized payload, exact window, static codebooks and calibration assets, allocation overhead, and protected K/V pairs with directory metadata. A storage implementation can remove any overlap between base and exact rows. Scoring $n$ candidates costs $O(n(d_K+d_V))$ arithmetic per head; a size-$N$ heap supports $O(n\log\max(2,N))$ selection; with $N=0$, retention can be skipped.

\end{document}